# Learning Deployable Navigation Policies at Kilometer Scale from a Single Traversal

**Jake Bruce**
QUT, Brisbane
jacob.bruce@hdr.qut.edu.au

**Niko Sünderhauf**
QUT, Brisbane
niko.suenderhauf@qut.edu.au

**Piotr Mirowski**
DeepMind, London
piotrmirowski@google.com

**Raia Hadsell**
DeepMind, London
raia@google.com

**Michael Milford**
QUT, Brisbane
michael.milford@qut.edu.au

**Abstract:** Model-free reinforcement learning has recently been shown to be effective at learning navigation policies from complex image input. However, these algorithms tend to require large amounts of interaction with the environment, which can be prohibitively costly to obtain on robots in the real world. We present an approach for efficiently learning goal-directed navigation policies on a mobile robot, from only a single coverage traversal of recorded data. The navigation agent learns an effective policy over a diverse action space in a large heterogeneous environment consisting of more than 2km of travel, through buildings and outdoor regions that collectively exhibit large variations in visual appearance, self-similarity, and connectivity. We compare pretrained visual encoders that enable precomputation of visual embeddings to achieve a throughput of tens of thousands of transitions per second at training time on a commodity desktop computer, allowing agents to learn from millions of trajectories of experience in a matter of hours. We propose multiple forms of computationally efficient stochastic augmentation to enable the learned policy to generalise beyond these precomputed embeddings, and demonstrate successful deployment of the learned policy on the real robot without fine tuning, despite environmental appearance differences at test time. The dataset and code required to reproduce these results and apply the technique to other datasets and robots is made publicly available at rl-navigation.github.io/deployable.

## 1 Introduction

Navigation is an important capability for many of the tasks to which robots could be applied. It is a fundamental skill for mobile intelligent agents, and requires decision-making across a diverse range of scales in time and space. Although motion primitives can be hard-coded at some level into mobile intelligent agents, navigating brand new environments requires some form of learning, and enabling machines to learn from experience has the potential to provide performance improvements across all areas of robotics. Unfortunately, interaction with the environment can be expensive for real robots, and reducing the amount of required real-world interaction is therefore a high priority for robot learning. Reinforcement learning (RL) algorithms have recently been shown to solve complex tasks including navigation with challenging input modalities such as vision, but these algorithms tend to require a large number of steps taken by the agent in the environment.

In this work, we train an agent to navigate to arbitrary goals in a large real-world environment, by interactively generating novel trajectories through a single coverage traversal of recorded data. We present an approach for the precomputation of visual representations, followed by on-the-fly stochastic augmentation of the finite recorded data for a computationally efficient training pipeline that can deliver more than 20,000 transitions per second of unique agent experience on a commodity desktop computer. We show that this technique scales to kilometer-scale heterogeneous environments with diverse action affordances, and we take robot navigation learning a significant step toward real world deployability by demonstrating successful policy execution on a real robot under natural variations in viewpoint and appearance. We freely publish the dataset and code for this work online at rl-navigation.github.io/deployable.

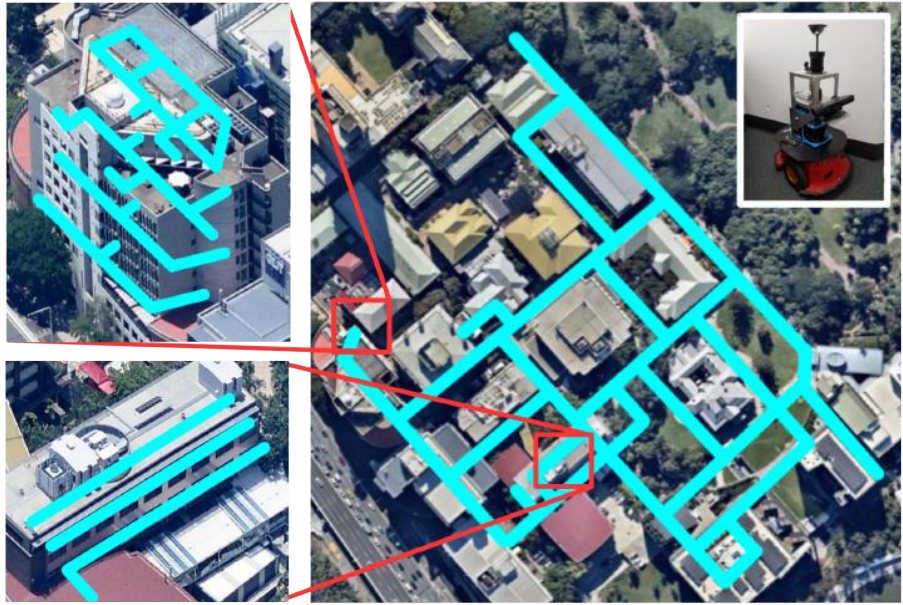

Figure 1: Graph of the environment built from a single traversal (inset showing robot), spanning the entire navigable area of the campus plus seven floors of two different buildings (zoomed-in areas). Note that the metric spatial arrangement of the graph shown here is not used in any way by the agent, and approximate metric information is used for visualisation purposes only.

## 2 Related Work

Robot navigation is a long-standing field of research, and a large diversity of approaches have been applied to the problems involved. Traditional approaches have consisted of localisation techniques, often in conjunction with mapping, in order to form as complete a picture as possible of the joint configuration of the world and the robot [1, 2]. Given such a joint configuration, a variety of traditional planning techniques can be applied to generate paths to desired goal locations. Alternative approaches consider mapless navigation, in which the agent navigates solely on the basis of its egocentric sensor modalities: mapping these inputs to actions with appearance-based matching and path recall [3], or by a process of implicit localisation and mapping in some form of latent representation [4, 5, 6]. In this work, we consider the mapless case, and learn to transform image input directly to actions without an explicit representation of the joint configuration of the robot and environment.

Research in the area of reinforcement learning has demonstrated remarkable success on tasks from playing games [7, 8] to continuous control of real [9] and simulated [10] robots. In the navigation domain, learning architectures have been proposed for creating explicit maps [11, 12], and agents have been trained to navigate without explicit maps using laser sensing [13, 14] and directly from images [15, 16, 17]. These existing approaches for goal-directed navigation consider only small spaces, or only virtual environments. [17] demonstrates large-scale learning from real-world imagery, but training and testing is done on the same environment, and also defines goals in terms of spatial coordinates or distances to landmarks. In contrast, our work trains a reinforcement learning agent to navigate to goal locations represented by an image of the goal, in an environment of kilometer scale, while also deploying the learned policy successfully on a real-world robot at test time.

However, end-to-end learning approaches still tend to suffer from a heavy data requirement. Often, agents can require upward of hundreds of millions of steps of interaction with the environment before converging to a competent policy. Approaches have been proposed for learning robot control in the real world from very little data [9], although it has proved difficult to scale these approaches to high-dimensional inputs and outputs. One approach for mitigating the real-world data requirement is to learn in a sufficiently high-fidelity simulation for a large number of environment steps before transferring the learned policy to the real world, with or without fine-tuning [18, 19]. High-quality simulations, especially simulations tuned to the specifics of the robot and environment in question, can be challenging to obtain. In contrast, our work requires no photorealistic rendering because we learn directly from real imagery.

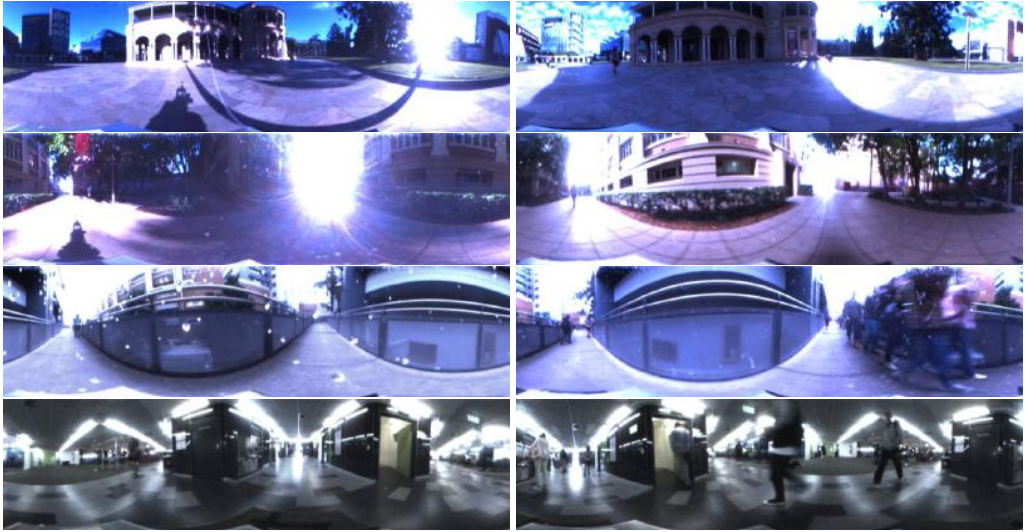

Figure 2: Example imagery from the omnidirectional camera, highlighting large differences between recorded training data (left) and input seen at deployment time (right). Differences include large illumination, shadow, and weather changes; image artifacts due to direct sunlight; raindrops on the camera lens; and people.

Models can be built from the real environment in which the agent is embodied in order to fit directly the specifics of the robot [20, 21, 22, 23]. In this work we learn from a single coverage traversal of the real world, gathered on the real robot, to remove the requirement for photorealistic simulation entirely while allowing the agent to generate the large number of novel trajectories it requires to learn to navigate [24]. This work is the first, to the best of our knowledge, to demonstrate successful deployment of kilometer-scale navigation policies trained by reinforcement learning on a real robot.

## 3 Approach

In this work, we propose to train a graph-navigation agent on data obtained from a single coverage traversal of its operational environment, and deploy the learned policy in a continuous environment on the real robot. We formulate the navigation problem as a partially observable Markov decision process (POMDP), described as a set of states $S$, a set of discrete actions $A$, a stochastic transition function $T(s, a) \to p(s'|s, a)$, a stochastic observation function $O(s) \to p(o|s)$, and a reward function $r(s, a)$. The goal of reinforcement learning is to obtain a policy $\pi(o) \to p(a|o)$ that generates actions that maximize total reward, potentially discounted by time. We implement a sparse reward structure, with a reward of $1.0$ delivered only when the agent reaches the precise goal location, so the policy that maximizes total discounted reward is the policy that follows the shortest path to the goal.

### 3.1 Interactive replay

We construct a simple world model from a single coverage traversal of the environment, allowing an RL agent to generate a large number of synthetic trajectories through real sensory inputs in an offline manner, without further interaction with the real environment during training. In this section, we describe the navigation graph and its construction, and the stochastic augmentations we perform in order to improve deployment robustness.

#### 3.1.1 Navigation graph

We build a navigation graph with a realistic small amount of human interaction: the user is responsible for annotating loop closures in the traversal, and eliminating duplicated regions of space. The result is a topological map of the environment (consisting of 2099 equally spaced nodes representing over 2km of traversable paths, see Figs. 1 and 3), each node being associated with an image in the recorded data, with edges between nodes representing mutual traversability. We use the omnidirectional camera on our robot platform to simulate rotational actions in the virtual agent, since it can see in all directions. Although loop closures and redundant places are detected with human interaction in this work, we emphasize that this procedure could be completely automated with techniques from Simultaneous localisation and Mapping (SLAM) and place recognition [2, 25].

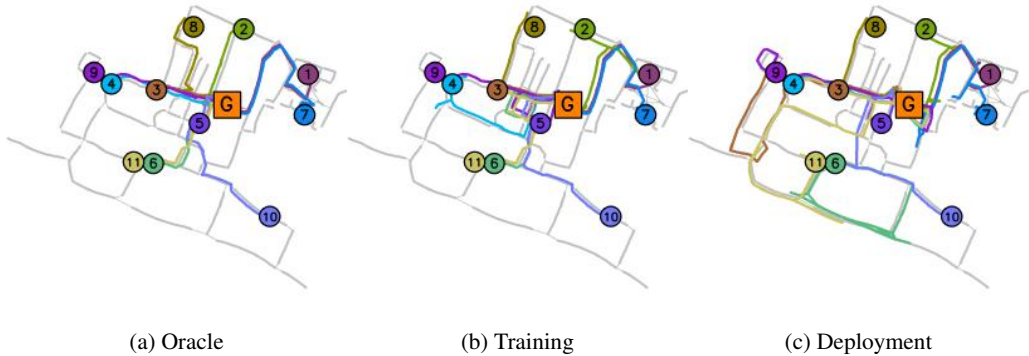

(a) Oracle  (b) Training  (c) Deployment

Figure 3: Trajectories from a variety of initial locations to a fixed goal using a) an optimal planner, b) a trained agent on the recorded training data, and c) the same trained agent navigating a robot through the real environment several weeks after the training data was recorded.

Our environment includes seven floors of two different buildings, and the elevators connecting these floors present an opportunity to evaluate learning a diverse action space. We encode basic motion by allowing the agent to turn left or right in 90° increments, and move forward by 1 meter. Elevators are a challenge however, because the simplest approach of allowing vertical motion in increments of single floors is an annoyance to humans sharing the elevator with the robot, and it would impose a burden on the elevator system if every user were to operate this way. We choose instead to provide a separate action for each possible destination floor, for a total of seven elevator actions in addition to the three basic movements. Destination-focused action spaces like this are common in human navigation modalities such as bus routes, and provide interesting challenges for robotics [26].

### 3.2 Stochastic environment

The concept of learning from the smallest possible set of recorded data is generally at odds with learning a robust, generalisable policy. In particular, a finite navigation graph can result in an agent simply memorising simplistic associations between inputs and actions that would not transfer to deployment. We address this problem by making it impossible for the agent to memorize the task: we inject multiple forms of nondeterminism in the observation and transition functions of our POMDP, which has been shown to be crucial to avoid overfitting on small amounts of navigation data [24].

#### 3.2.1 Nondeterministic frame selection

Although the navigation graph consists of discrete locations separated by 1 meter, the recorded data includes all camera images produced between each of these locations. This amounts to approximately 30 extra viewpoints per discrete location, given our 15-Hz camera on a robot moving at 0.5 meters per second. We use these additional images as a pool of potential observations associated with each location in the graph, to provide variations of the sort an agent might experience at deployment time. When queried for an observation given a state, our navigation environment samples an image captured within 0.5m of that state, followed by a small rotation of up to 8 degrees in either direction, to provide variation in inputs and mitigate differences between poses observed at training and test time. In our graph of 2099 unique locations, each discrete state is augmented with an average of 120 of these alternative observations, bringing the total number of real observations the agent can experience to over 250,000. Although this is a diverse set of possible observations, an expressive function approximator could potentially memorise the entire set; for this reason, we also augment the images on the fly with unique noise described below.

### 3.2.2 Input noise

We augment the agent's observations on the fly with two forms of unique noise to account for differences in pixel intensities upon deployment. Changes in brightness and camera exposure are common between training and test time imagery, and these changes are correlated in time: videos do not generally exhibit uncorrelated flicker. To emulate these gradual shifts in apparent brightness, we add zero-mean temporally correlated Ornstein-Uhlenbeck noise [27] denoted by $\xi_x$, parameterized by $\sigma_{\text{global}}$ and $\theta_{\text{global}}$ to the input space. However, due to the nonlinear transformation performed by the pretrained visual encoder (see Section 3.3.2) uniform shifts in image brightness do not correspond to uniform shifts in feature space. To simulate input noise in feature space, we approximate the propagation of uniform brightness shifts through the visual encoder with a first-order Taylor approximation $\xi_{\phi_j} = \xi_{x_i} J_{ij}$ where $J_{ij}$ is the partial derivative of feature $\phi_j$ with respect to a global additive shift in brightness applied to image $x_i$. To further mitigate overfitting and differences at deployment time, we also add a small amount of uncorrelated zero-mean Gaussian noise to the feature vectors parameterised by $\sigma_{\text{local}}$, scaled by the magnitude of $J_{ij}$. Combining the two forms of noise results in a correlated global change in features due to brightness, plus an uncorrelated shift for each feature scaled by its sensitivity to the input.

Because we augment the input in feature space rather than in image space, we can apply this noise to the precomputed embeddings and avoid running the relatively expensive forward pass of the visual encoder, allowing more computation to be dedicated to the policy itself. Through this technique we are able to achieve more than 20K transitions per second on a single commodity desktop computer. These modifications are simple, but surprisingly effective and computationally efficient ways of providing unlimited variability to the observations of the agent, and stochastic augmentations like these have been shown to be crucial for achieving generalisation beyond the training data [24].

### 3.2.3 Stochastic transitions

Since the environment consists of a fixed graph, successful policies could be learned that consist of simply associating each initial observation-goal pair with a memorized sequence which is then executed open-loop. These degenerate policies would not cope well with deployment-time differences in execution such as precise distances and rotations between nodes. To mitigate this issue, we use a stochastic transition function in our POMDP [28]: each *move forward* action has a probability $p_{\text{stutter}}$ of leaving the agent in the same state rather than transitioning forward.

## 3.3 Training

We formulate the navigation problem as a sparse-reward POMDP in which the agent is spawned randomly at the beginning of each episode at any location in the graph, and a reward of 1.0 is delivered only upon reaching a curriculum-determined goal. The goal is defined as a precise location in the graph, independent of agent orientation, and is presented to the agent in the form of an image $g_t$. To train the agent we use a popular deep reinforcement learning algorithm called A3C [29], which has been applied successfully to challenging navigation tasks with image input [15, 17]. The goal-directed A3C architecture in this work directly optimizes a conditional distribution over actions $\pi(a_t|o_t, g_t)$ by estimating the gradient of expected return from short Monte Carlo rollouts of experience, bootstrapped by an estimate of future reward given by the value function $V(o_t, g_t)$. A3C parameterises both the policy and value functions as deep neural networks in order to learn directly from complex inputs such as images. This section describes our curriculum training strategy and the architecture of the reinforcement learning agent. All hyperparameters are given in Tab. 2.

### 3.3.1 Curriculum learning

In tasks with very sparse rewards such as ours, where rewards are only ever delivered upon reaching the goal, agents are unlikely to observe many rewards early in training. We mitigate this problem by training the agent on a curriculum of gradually increasing difficulty [30, 31, 17]. We construct a curriculum of $N_c$ levels, each level $i$ consisting of all pairs of locations that are connected by a path of less than or equal to $\frac{i}{N_c} L_{\text{max}}$, where $L_{\text{max}}$ is the longest optimal path in the graph. We gradually increase the difficulty of the task in a manner that adapts to the learning progress of the agent by incrementing the curriculum level whenever the agent reaches a success rate of 80% as in [31]. Using this curriculum, the agent starts training on paths no longer than 5 meters and gradually progresses up to 403 meters, the longest optimal path in our graph.

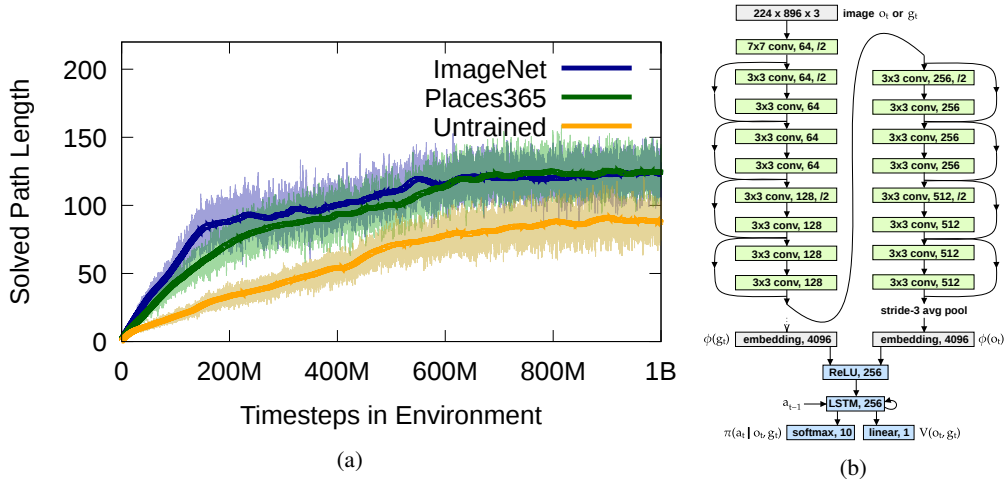

Figure 4: a) Learning curve for three different visual encoders, trained for object classification on ImageNet, place categorisation on Places365, and a randomly initialised network. The performance is measured by the optimal path length of the tasks solved by the agent, where failed episodes are assigned a value of zero. b) Architecture of the agent: fixed encoder module in green, and learned policy in blue.

### 3.3.2 Architecture

The network architecture consists of a pretrained and fixed ResNet18 encoder [32] shown in Fig. 4 which processes the observation and goal images $o_t$ and $g_t$, followed by strided mean pooling at the final feature layer resulting in image embeddings $\phi(x)$ of length 4096 per image. Pretrained and fixed vision processing allows us to preprocess all of the images in our dataset, enabling a large gain in computational efficiency over running the encoder every frame, and fixing the visual encoder has been shown to reduce overfitting to small environments [24]. We evaluate three different visual encoders and report their learning curves in Fig. 4: ImageNet [33] pretraining, Places365 [34] pretraining, and random initialisation. ImageNet and Places365 networks achieve comparable performance, so we choose Places365 weights for deployment. The task of condition-invariant place categorisation may provide features more suitable for navigation under the changing conditions of deployment, although we have not explicitly evaluated this.

The representations of the observation and goal are concatenated and used as the input for an architecture used in [17]: a fully-connected layer of width 256 followed by ReLU, which is then concatenated with a one-hot representation of the previous action $a_{t-1}$ and passed to an LSTM [35] of width 256. Recurrence in the architecture allows the agent to accumulate localisation evidence over multiple timesteps, which has been shown to be a useful capability for navigation [15]. The LSTM representation is then projected to a softmax layer of width $N_a$ parameterising the categorical distribution $\pi(a_t|o_t, g_t)$, and a single linear unit to predict $V(o_t, g_t)$.

## 4 Evaluation

We evaluate the ability of our agent to transfer knowledge from the recorded data to the real world, by deploying the learned policy on the same Pioneer 3DX mobile robot, shown in Fig. 1. The deployment test was conducted three weeks after the initial training data was collected, at different times of day (with evaluations spanning a total of approximately 8 hours and nearly 5km of navigation), to allow for a reasonable amount of natural variation between training and test conditions. We consider a goal-directed navigation task, navigating to a fixed goal in the center of the campus from a diverse set of initial locations in the environment. We choose this task for clarity of interpretation and evaluation, although the agent has been trained not just on fixed goals, but every pair of start and goal locations in the graph. The goal location was chosen to maximize the density of nearby nodes to encourage the agent to exhibit a diversity of paths for analysis.

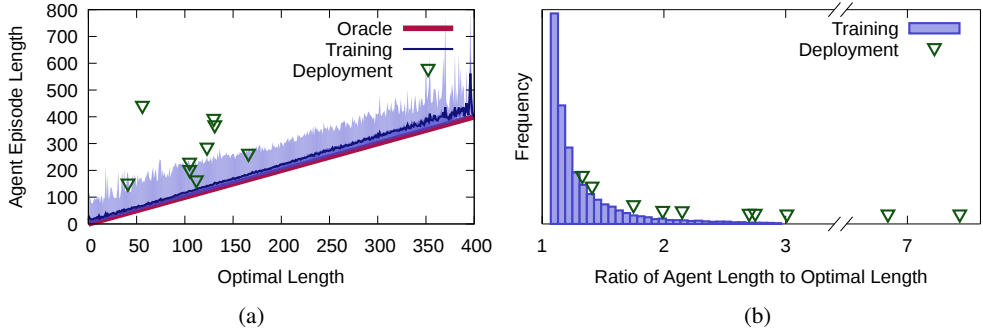

(a)          (b)

Figure 5: a) Distance travelled to the goal as a function of shortest path distance. Performance on the training data is measured over 100K episodes on the trained agent with random start and goal locations. Deployment trajectories were gathered at different times of day several weeks after the training data was recorded. The middle $50^{th}$ and $95^{th}$ percentiles are shaded dark and light blue, respectively. b) Histogram of the ratio of the agent's path length to the optimal path length. The agent generates paths that are 1.35x longer than the optimal path on average. Deployment paths are longer than their virtual counterparts due to perceptual differences, but roughly follow the distribution.

| | | | Trajectory Length (in meters) | | | | | | |
|---|---|---|---|---|---|---|---|---|---|
| # | Oracle | Training | Deployment | Ratio | # | Oracle | Training | Deployment | Ratio |
| 1 | 139 | 147 | 375.0 | 2.70 | 7 | 138 | 157 | 379.0 | 2.75 |
| 2 | 113 | 133 | 225.3 | 1.99 | 8 | 120 | 133 | 159.4 | 1.33 |
| 3 | 64 | 78 | 437.6 | 6.84 | 9 | 131 | 136 | 281.7 | 2.15 |
| 4 | 113 | 118 | 197.8 | 1.75 | 10 | 184 | 209 | 258.8 | 1.41 |
| 5 | 49 | 68 | 147.4 | 3.01 | 11 | 120 | 142 | 891.3 | 7.43 |
| 6 | 106 | 121 | — | — | Avg | | | | 2.42 |

Table 1: Distances travelled during navigation experiments on the real robot. Trajectory 6 failed to reach the goal after travelling more than 1km. In the rightmost column we report the ratio of the path lengths in the real world to the optimal path length emitted by the oracle. The agent sometimes takes significantly longer paths than the optimal planner due to navigation errors, caused by perceptual differences between training and test time.

### 4.1 Robot control

Due to the nature of the training data as a fixed traversal, our agent learns high-level navigation actions at a level of abstraction similar to navigation instructions a human might convey, rather than direct low-level control: commands of the form "turn left", "go straight", or "take the elevator to the 4th floor". In order to evaluate on a real robot a policy composed of these high-level actions, we convey the selected action to a human operator responsible for carrying out the action.

The agent is trained solely on image input, but for safety the robot is equipped with a scanning laser rangefinder that allows it to detect obstacles. When a movement action is emitted by the agent, but the movement would collide with an obstacle, we instead resample a new action from the policy without replacement. We apply the same procedure in the case of obstacles unseen by the laser: transparent and reflective obstacles, fast-moving dynamic objects, unobservable safety issues like fall hazards, and roads with vehicle traffic, which are not part of the training data.

### 4.2 Results

We evaluate the agent on a goal-directed navigation task of reaching the center of the campus from a diverse set of eleven initial locations, with results shown in Figs. 3 and 5 and Tab. 1. The agent navigated successfully to the goal, defined by arriving within 2 meters of the location where the goal image was captured, in ten of the eleven trials. On the failed trajectory 6, the agent repeatedly navigated back and forth along the same path for 1km and never reached the goal. Notably, the failed path was one of the least-used routes (as measured by shortest-path visitation) due to its location on the periphery, and also exhibited large differences due to shadows and sun glare between the afternoon appearance at training time and morning appearance at test time. These factors are likely to be responsible for the poor deployment performance in this region and point to interesting potential improvements that can be made to our approach, such as extending our feature-space augmentation strategy to simulate the effect of sun glare and shadows in feature space, and adapting the curriculum learning strategy to ensure equal coverage of the graph during training.

In terms of path length, the deployment trajectories were longer than their virtual counterparts on the training data, traveling 2.42 times farther than the optimal path on average, compared to 1.14 times farther for the virtual agent on these paths. This is not surprising in the absence of perfect localisation, since perceptual differences of the sort that caused trajectory 6 to fail can cause less severe navigation errors that inevitably lengthen the path, even when followed by successful recovery. Empirically, as can be seen in videos of the experiments at rl-navigation.github.io/deployable, the agent often doubles back on its path one or more times before recovering. When compared against the distribution of path lengths shown in Fig. 5 with the exception of notable outlier trajectories 3 and 11, the deployment path lengths are reasonably close to the distribution exhibited by the virtual agent.

| Hyperparameter | Value | Hyperparameter | Value |
| --- | --- | --- | --- |
| Optimiser | Adam | # of discrete actions $N_a$ | 10 |
| Learning rate | $10^{-4}$ | # of random rotations per image | 5 |
| Reward discount $\gamma$ | 0.99 | Maximum stochastic rotation | 8° |
| Entropy bonus weight | $5 \cdot 10^{-4}$ | Correlated noise $\sigma_{\text{global}}$ | 0.01 |
| A3C workers | 128 | Correlated noise $\theta_{\text{global}}$ | 0.15 |
| Monte Carlo rollout length | 50 | Uncorrelated noise $\sigma_{\text{local}}$ | 0.01 |
| Preprocessing stride | 3 | Transition stutter $p_{\text{stutter}}$ | 0.05 |
| Distance between graph nodes | 1 meter | Curriculum levels $N_c$ | 100 |
| Granularity of rotation action | 90° | Curriculum increment threshold | 0.8 |

Table 2: Hyperparameters used for training and deployment experiments.

## 5 Discussion

We have demonstrated an approach for learning navigation policies in large heterogeneous environments from a single traversal, exploiting powerful pretrained visual feature extraction and a data augmentation procedure involving nondeterministic frame selection across position and orientation, and efficient local and global feature-space noise that enables throughput of over 20K transitions per second on a commodity desktop computer. Allowing an agent to generate millions of novel trajectories through a set of recorded data enables the magnitude of experience required by model-free reinforcement learning systems, without requiring a photorealistic simulator, resulting in policies that successfully execute high-level navigation behavior on the real robot at deployment time without fine tuning. We make publicly available the dataset and code required to reproduce these results, and to apply the technique to other datasets and robots, at rl-navigation.github.io/deployable.

Gathering a coverage traversal of the environment by teleoperation requires more human interaction than would a robot learning purely on its own, although fully autonomous data collection could be achieved if required. In practice, however, for safety most robot experiments are heavily supervised by human operators. Since our approach requires only a single pass through the environment and minimal annotation afterward, producing a navigation graph requires no more human intervention than does a typical SLAM system.

The simple yet effective stochastic observation mechanism we apply in this work achieves reasonable deployment-time performance, but further minimising perceptual differences between training and test input distributions will help the deployed agent approach the performance it exhibits in simulation. Neural network features like we use here have shown promising results toward robustly recognising places in the face of appearance and viewpoint change [36]; however, reducing perceptual aliasing and appearance differences is an active area of research in place recognition and SLAM [25] and many other techniques developed for that purpose may also provide improvements in work like ours.

Future work will focus on integrating low-level behavior for a fully autonomous system. If approximate localisation is available, then movement throughout the graph can be handled by waypoint control; however, contributions in visual servoing [37] and teach and repeat [38] demonstrate that a purely visual solution is also possible.

**Acknowledgments**

We wish to thank colleagues Adam Jacobson, Gavin Suddrey, Vibhavari Dasagi, Robert Lee, Serena Mou, William Hooper, Dominic Jack and William Chamberlain for technical assistance and valuable discussion. This work was funded in part by the National Science and Engineering Research Council of Canada, and the Australian Research Council Centre of Excellence for Robotic Vision (project number CE140100016). Michael Milford is supported by an Australian Research Council Future Fellowship (FT140101229).

# Supplementary Material for "Learning Deployable Navigation Policies at Kilometer Scale from a Single Traversal"

## 6 Navigation Graph

In this section we present information about our particular navigation graph. The graph spans an entire university campus, with 2099 nodes separated by approximately 1 meter. The graph contains 7 different floors of two different buildings, so some of the edges in the graph correspond to elevator travel. Nodes within 4 meters of an elevator are connected to each other accessible floor in the same building.

Fig. S6a shows the navigation graph in the form of an inflated heatmap, where each node is colored and inflated in size to represent how often it participates in an optimal path. Some nodes, especially at intersections and near the ground floor of the building in the top right, participate in many more optimal paths than others.

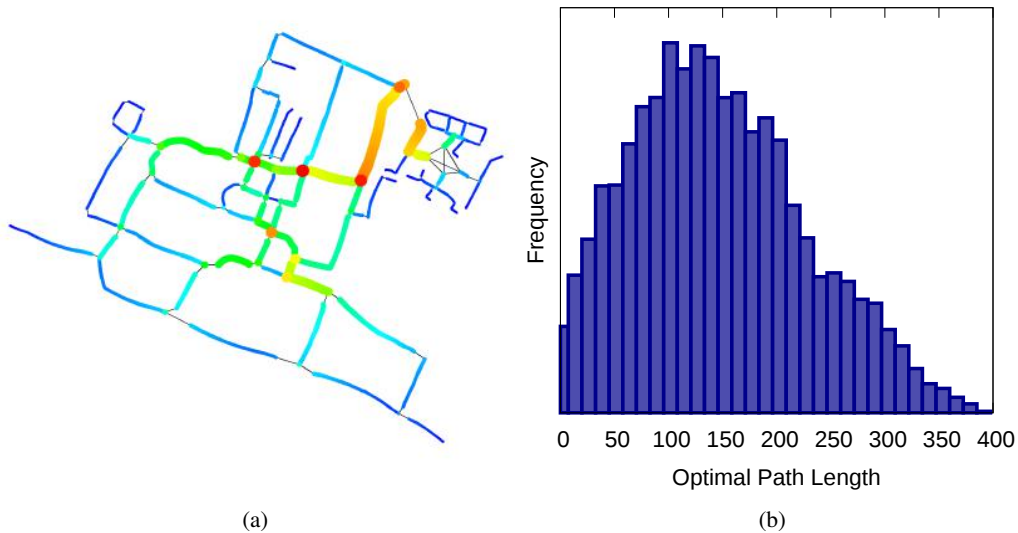

(a)          (b)

Figure S6: a) Heatmap of the visitation count of each node on all pairs of shortest paths in the graph. Nodes are colored and inflated to represent their participation counts across shortest paths; this visualises the bottlenecks in the graph, which represent regions that participate in a large number of optimal paths. b) Histogram of optimal path lengths in the graph over all pairs of start and end nodes, with a mean of 152.6m and standard deviation of 80.7m.

# 7  Localisation Estimate

In addition to the navigation policy, we trained a single-layer softmax classifier to predict the current location of the agent, given the recurrent state of the policy LSTM. We inserted a stop-gradient operation for this readout layer: its gradients were not used to train the navigation agent. The readout is only a single linear projection and the gradients did not propagate into the rest of the network, so the localisation layer reached a top-1 accuracy of only approximately 60%. Analysis of its activity however can provide some insight into the degree of localisation information that is represented in the activations of the policy network, which was trained only to maximise reward. Fig. S7 shows a heatmap of the log-probability of the localisation readout at a location in the training data and a similar location at deployment time. Deployment-time localisation distributions are much less crisp, which may explain the performance decrease of the transferred policy.

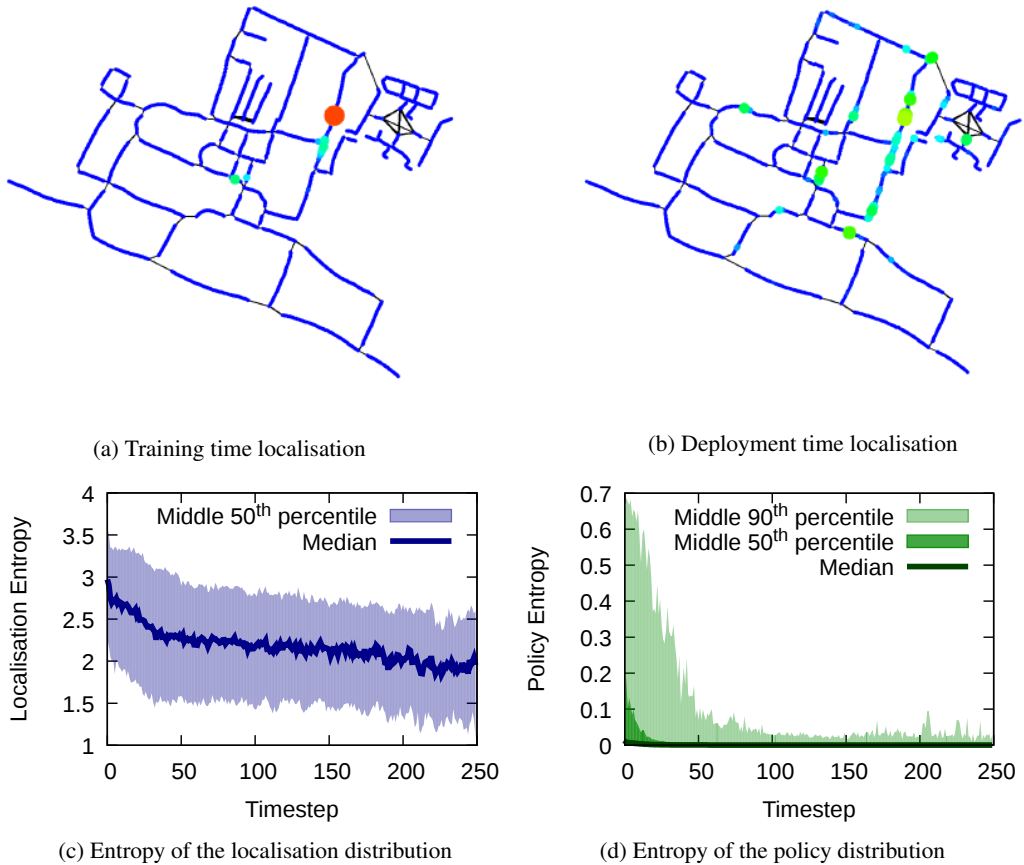

(a) Training time localisation

(b) Deployment time localisation

(c) Entropy of the localisation distribution

(d) Entropy of the policy distribution

Figure S7: a,b) Heatmap of the log-probability of the localisation readout layer for each node in the graph, when the agent was at nearly the same location at training and deployment time during trajectory 1. As shown in the video available at `rl-navigation.github.io/deployable`, the agent doubled back and repeated this section of the graph several times before recovering and eventually reaching the goal. This discrepancy between localisation information at training at deployment time, ultimately caused by perceptual differences such as the presence of humans and raindrops on the camera lens, may explain the navigation errors made by the deployed agent. c,d) Entropy of the localisation and policy distributions, plotted against the timestep of the episode, measured over 200K timesteps of the trained agent in the training environment. The entropy of both distributions decreases over time at the beginning of the episode; this indicates that the recurrent network accumulates observations and becomes more certain over time as intended.

## 8 Deployment Trajectories

Each deployment trajectory is presented individually in Figs. S8-S18 for ease of interpretation. Initial locations are indicated by a circle (color-coded to match Fig. 3) and the goal location is represented by the orange square labelled **G**. The goal location is the same for each traversal, and was chosen to maximize the diversity of optimal routes from different initial locations, encouraging a wide variety of trajectories during deployment. Most trajectories are qualitatively similar between the oracle, training data, and real world, although slightly different routes are common, and at deployment time the agent often doubles back or takes a short wrong turn before recovering. Videos of every trajectory are available at `rl-navigation.github.io/deployable`.

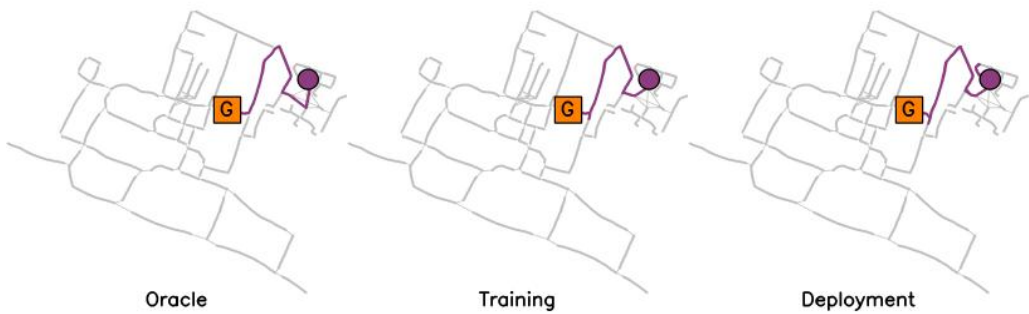

Figure S8: Trajectory 1 (successful)

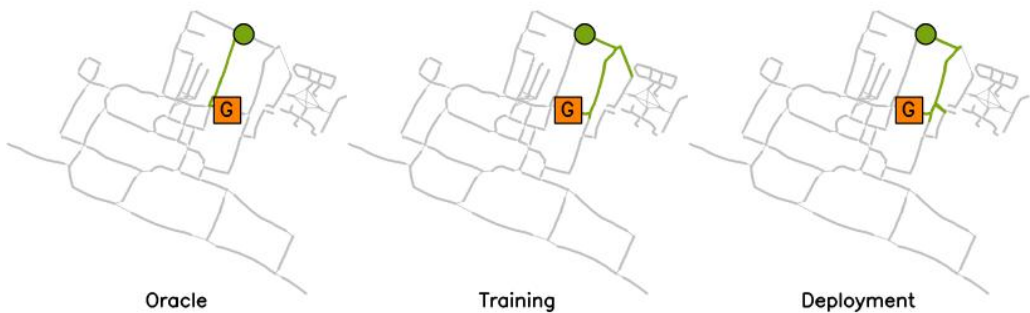

Figure S9: Trajectory 2 (successful)

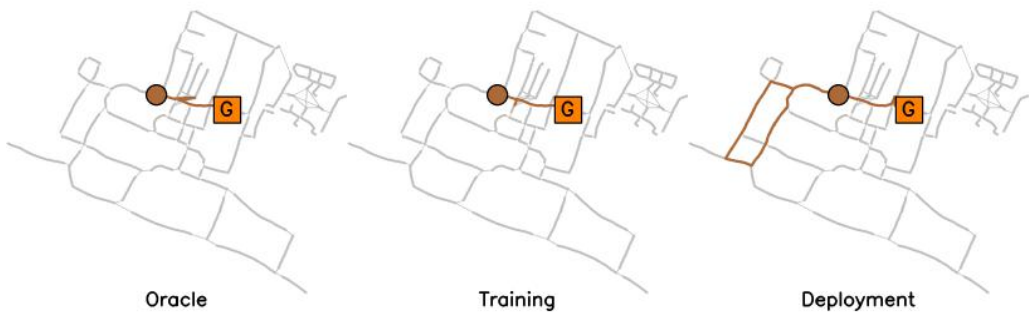

Figure S10: Trajectory 3 (successful)

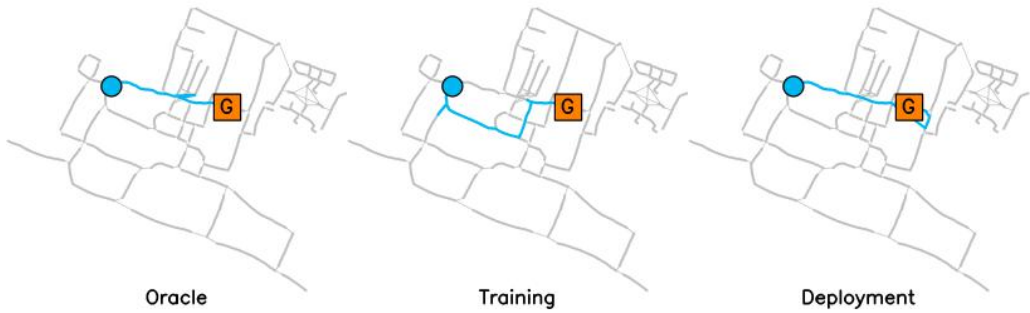

Figure S11: Trajectory 4 (successful)

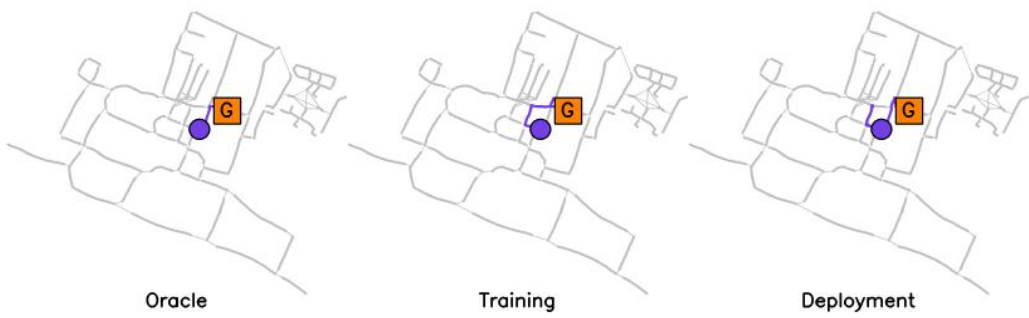

Figure S12: Trajectory 5 (successful)

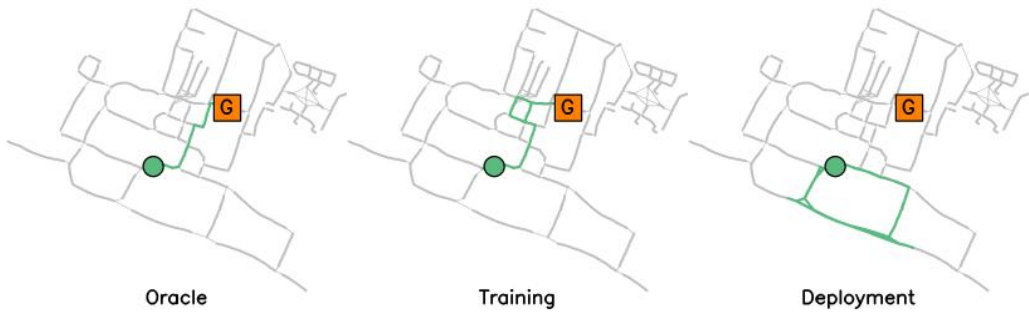

Figure S13: Trajectory 6 (failed)

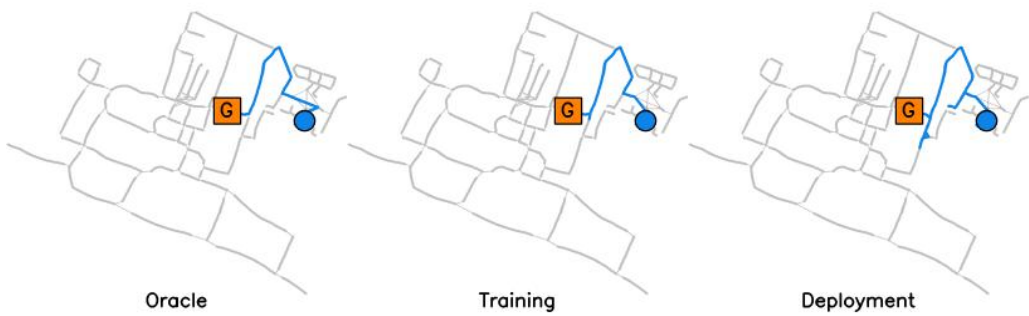

Figure S14: Trajectory 7 (successful)

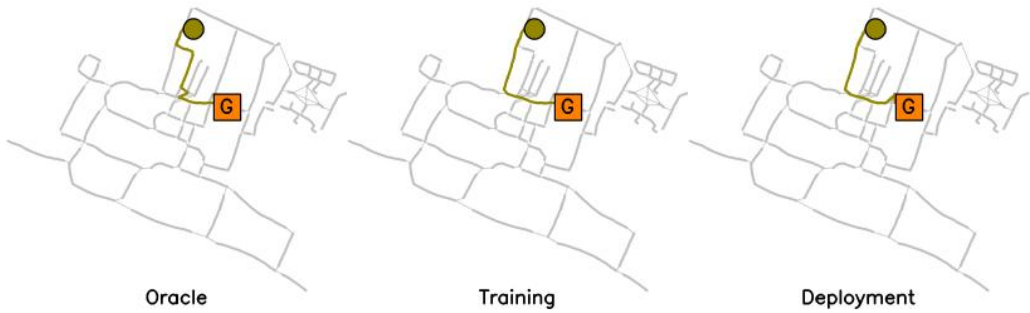

Figure S15: Trajectory 8 (successful)

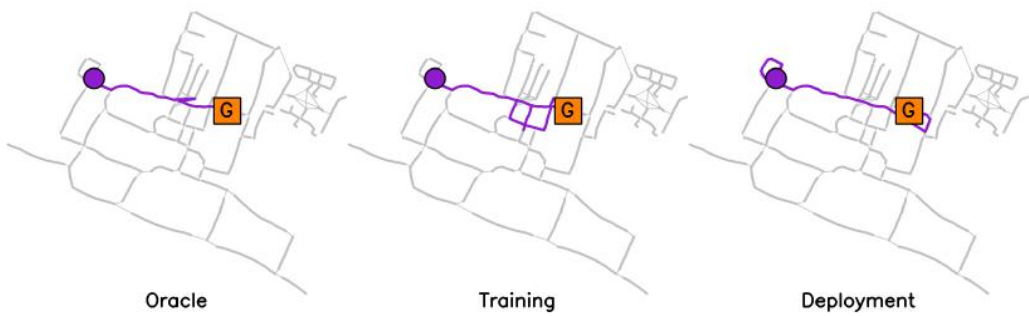

Figure S16: Trajectory 9 (successful)

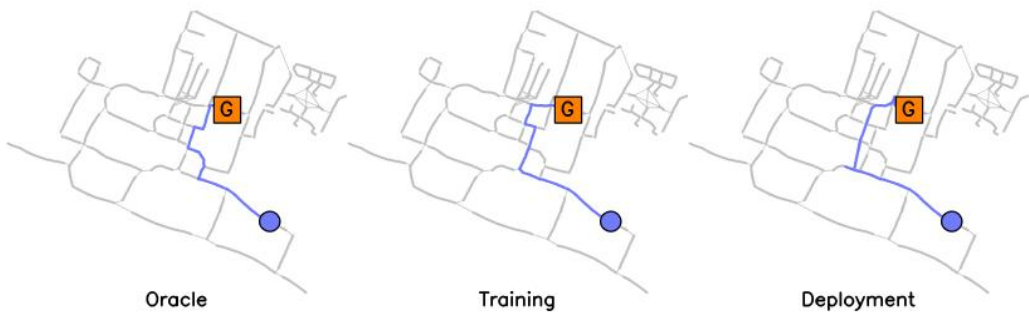

Figure S17: Trajectory 10 (successful)

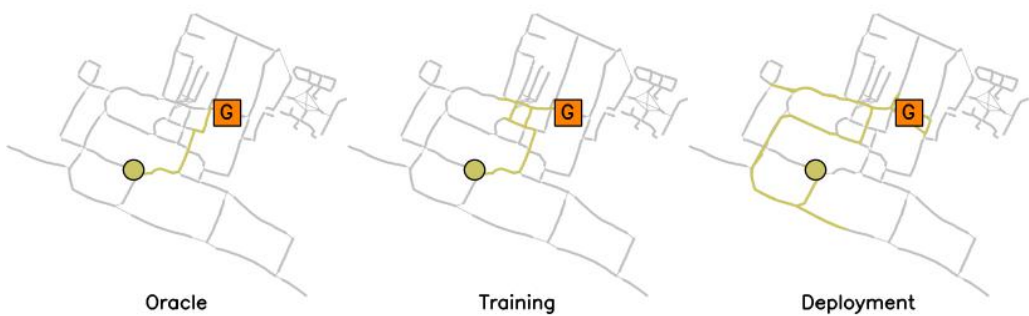

Figure S18: Trajectory 11 (successful)

15